\documentclass{article} % For LaTeX2e
\usepackage{iclr2024_conference,times}

\usepackage{amsmath,amsfonts,bm}

\def\eqref#1{equation~\ref{#1}}
\def\1{\bm{1}}

\DeclareMathAlphabet{\mathsfit}{\encodingdefault}{\sfdefault}{m}{sl}
\SetMathAlphabet{\mathsfit}{bold}{\encodingdefault}{\sfdefault}{bx}{n}

\usepackage{booktabs}
\usepackage{sansmath}
\usepackage{mathtools}
\usepackage{graphicx}
\usepackage{caption}
\usepackage{subcaption}

\usepackage{cite}
\usepackage{amsmath, amsfonts, amssymb}
\usepackage{algorithmic}
\usepackage{graphicx}
\usepackage{textcomp}

\usepackage{amsthm}
\newtheorem{example}{Example}

\newtheorem{lemma}{Lemma}
\newtheorem{definition}{Definition}
\newtheorem{corollary}{Corollary}
\newtheorem{remarks}{Remarks}

\usepackage{hyperref}
\usepackage{url}

\title{Activations Through Extensions: \\ A Framework to Boost Performance of Neural Networks}

\author{Chandramouli Kamanchi$^{ \ast}$\\
IBM Research\\
Bangalore, India \\
\small{\texttt{chandramouli.kamanchi@ibm.com}}
\And
Sumanta Mukherjee\thanks{Equal Contribution}\\
IBM Research\\
Bangalore, India\\
\small{\texttt{sumanm03@in.ibm.com}}
\And
Kameshwaran Sampath\\
IBM Research\\
Bangalore, India\\
\small{\texttt{kameshwaran.s@in.ibm.com}}
\And
Pankaj Dayama\\
IBM Research\\
Bangalore, India\\
\small{\texttt{pankaj.dayama@in.ibm.com}}
\And
Arindam Jati \\
IBM Research\\
Bangalore, India\\
\small{\texttt{arindam.jati@ibm.com}}
\And
Vijay Ekambaram \\
IBM Research\\
Bangalore, India\\
\small{\texttt{vijaye12@in.ibm.com}}
\And
Dzung Phan \\
IBM T.J. Watson Research Center, \\
New York \\
\small{\texttt{phandu@us.ibm.com}}
}

\iclrfinalcopy % Uncomment for camera-ready version, but NOT for submission.
\begin{document}

\maketitle

% 
% \begin{abstract}
% These instructions give you guidelines for preparing papers for 
% IEEE Transactions and Journals. Use this document as a template if you are 
% using \LaTeX. Otherwise, use this document as an 
% instruction set. The electronic file of your paper will be formatted further 
% at IEEE. Paper titles should be written in uppercase and lowercase letters, 
% not all uppercase. Avoid writing long formulas with subscripts in the title; 
% short formulas that identify the elements are fine (e.g., "Nd--Fe--B"). Do 
% not write ``(Invited)'' in the title. Full names of authors are preferred in 
% the author field, but are not required. Put a space between authors' 
% initials. The abstract must be a concise yet comprehensive reflection of 
% what is in your article. In particular, the abstract must be self-contained, 
% without abbreviations, footnotes, or references. It should be a microcosm of 
% the full article. The abstract must be between 150--250 words. Be sure that 
% you adhere to these limits; otherwise, you will need to edit your abstract 
% accordingly. The abstract must be written as one paragraph, and should not 
% contain displayed mathematical equations or tabular material. The abstract 
% should include three or four different keywords or phrases, as this will 
% help readers to find it. It is important to avoid over-repetition of such 
% phrases as this can result in a page being rejected by search engines. 
% Ensure that your abstract reads well and is grammatically correct.
% \end{abstract}
% 

\begin{abstract}
  % A clear and well-documented \LaTeX\ document is presented as an
  % article formatted for publication by ACM in a conference proceedings
  % or journal publication. Based on the ``acmart'' document class, this
  % article presents and explains many of the common variations, as well
  % as many of the formatting elements an author may use in the
  % preparation of the documentation of their work.

% Activation functions play a pivotal role in neural networks by introducing non-linearities and allowing them to learn complex mapping between inputs and outputs. 
% In mathematics literature, a function ${\displaystyle F}$ is called as an extension of another function ${\displaystyle f}$ if whenever ${\displaystyle x}$ is in the domain of ${\displaystyle f}$ then ${\displaystyle x}$ is also in the domain of ${\displaystyle F}$ and ${\displaystyle f(x)=F(x).}$

Activation functions are non-linearities in neural networks that allow them to learn complex mapping between inputs and outputs. Typical choices for activation functions are ReLU, Tanh, Sigmoid etc., where the choice generally depends on the application domain.
In this work, we propose a framework/strategy that unifies several works on activation functions and theoretically explains the performance benefits of these works. We also propose novel techniques that originate from the framework and allow us to obtain ``extensions'' (i.e. special generalizations of a given neural network) of neural networks through operations on activation functions. We theoretically and empirically show that ``extensions'' of neural networks have performance benefits compared to vanilla neural networks with insignificant space and time complexity costs on standard test functions. We also show the benefits of neural network ``extensions'' in the time-series domain on real-world datasets.

\end{abstract}

\section{Introduction}
Current literature on machine learning and artificial intelligence is filled with triumphs of neural networks over other alternatives. A significant portion of these successes could be attributed to the design of novel and innovative activation functions. Here we comment on articles that are relevant to our work.

ReLU, Sigmoid, and Tanh are arguably the most common activation functions of choice in artificial neural networks of which ReLU is more predominant for its simplicity and computational efficiency. Several enhancements and modifications for these activation functions are proposed in the literature. For example \citet{maas2013rectifier} proposes Leaky ReLU and shows performance improvement over ReLU on benchmark tasks in the acoustics domain. 
In \citet{he2015delving}, authors come up with a parametrization of ReLU (PReLU). It is shown that PReLU, along with a novel initialization procedure, surpasses human-level performance on the ImageNet classification dataset. Of late, \citet{maniatopoulos2021learnable} introduces \textbf{Le}arnable \textbf{Le}aky Re\textbf{LU} (LeLeLU), which is a further parameterized variant of Leaky ReLU.

In \citet{46503}, authors describe exhaustive search and reinforcement learning paradigm based techniques to find activation functions and discover the now well-known Swish activation function and show that it surpasses the performance of ReLU on several tasks. The Swish activation could be seen as a modification of Sigmoid. Similarly, \citet{biswas2021tanhsoft} proposes several variants of Tanh by taking advantage of parametrization.

In recent times, custom activation functions have been explored in the backdrop of transformer based architectures as well. For example \citet{fang2022transformers} investigates rational activation functions for transformer based architectures and shows their effectiveness on the \textbf{G}eneral \textbf{L}anguage \textbf{U}nderstanding \textbf{E}valuation (GLUE) benchmark. Their choice of activation function is very similar to \citet{molina2019pad}, where the latter calls them \textbf{P}ad\'e \textbf{A}ctivation \textbf{U}nits (PAUs).

Many of the novel activation functions in the literature are parameterized splines. For example \citet{tavakoli2021splash} introduces an activation function called SPLASH, an acronym for \textbf{S}imple \textbf{P}iece-wise \textbf{L}inear and \textbf{A}daptive with \textbf{S}ymmetric \textbf{H}inges, that is mathematically a spline. It has been shown that these SPLASH units outperform the traditional ReLU and its variants on MNIST, CIFAR-10 and CIFAR-100 benchmarks.

Other examples of activation functions in the literature that come from the spline family of functions are presented in \citet{jin2016deep, zhou2021learning}. While \citet{jin2016deep} proposes an S-shaped spline and calls it SReLU  \citet{zhou2021learning} proposes an activation function similar to \citet{tavakoli2021splash} and calls it PWLU, an acronym for \textbf{P}iece-\textbf{w}ise \textbf{L}inear \textbf{U}nit. Both variants are shown to outperform alternatives on standard benchmarks.

Polynomials have also been explored as activation functions. For example \citet{goyal2019learning} chooses a linear combination of multivariate monomials as an activation function and learns the coefficients of the linear combination.

Other approaches for novel activation functions is through evolutionary algorithms. For example \citet{bingham2022discovering} proposes a method called PANGAEA, an acronym for \textbf{P}arametric \textbf{A}ctivatio\textbf{N} functions \textbf{G}enerated \textbf{A}utomatically by an \textbf{E}volutionary \textbf{A}lgorithm. Essentially the method discovers architecture specific activation functions through evolutionary algorithms and optimizes the parameters of the so discovered activation functions through gradient descent.

Further additions to activation function literature involve combining  activation functions in a novel way. For example \citet{ma2021activate} introduces ACON, an acronym for \textbf{Ac}tivate \textbf{O}r \textbf{N}ot, that automatically allows a neural network to switch on/off an activation function at a given node of the neural network.

Authors of \citet{manessi2018learning} investigate convex and affine combinations of traditional activation functions and show improved performance on well known architectures namely LeNet-5, AlexNet and ResNet-56 compared to standard activation functions like ReLU, Tanh etc. Another work with a similar idea is \citet{apicella2019simple}, where they investigate a slightly different affine combination of activation functions that they call \textbf{V}ariable \textbf{A}ctivation \textbf{F}unction (VAF) subnetwork. \citet{sutfeld2020adaptive, liang2021reproducing} are other works in literature that investigate a combination of elementary activation functions and are very similar to these works. 

It is theoretically shown in \citet{du2018power} that over-parameterization of  neural networks enables global optimization and generalization albeit for the case of quadratic activation functions. Similar theoretical reasons could potentially be attributed to the success of so far considered over-parameterized activation functions in the literature.

Quite a few works in the literature choose task specific activation functions. For example \citet{ziyin2020neural} demonstrates the ineffectiveness of neural networks to model periodicity in the data and proposes a custom activation function to mitigate the problem.

Other additions to the literature on activation functions include merging activation functions with other layers of the neural network. For example \citet{liu2020evolving} combines Batch Normalization with activation functions in a novel way. A comprehensive survey of different activation functions in the literature is presented in \citet{dubey2022activation, kunc2024three}.

In this work, we propose a framework that allows construction of novel activation functions through mathematical extensions. Our contributions are 
\begin{itemize}
    \item We propose a framework based on extensions that unifies and theoretically explains the performance improvement of several activation functions proposed in the literature.
    \item We theoretically show that extensions are finer fit to the data.
    \item We propose novel learnable activation functions that originate from this framework.
    \item We demonstrate the effectiveness of these learnable activation functions on synthetic benchmark functions.
    \item We also demonstrate its effectiveness on real-world time series datasets.
\end{itemize}

\section{Background}
In this section we introduce notation and review feedforward neural networks and statistical learning theory very briefly.
\subsubsection{Feedforward Neural Networks}
A vanilla feedforward neural network \citet{haykin2009neural} (FNN) consists of $L$ layers. Each layer has $M_l, 1 \leq l \leq  L$ hidden nodes. Given an input vector $x[1\colon t] = [x_1, x_2, \cdots, x_t]'$, the output $\hat{y}$ of FNN is obtained as follows.
In the first layer, we construct pre-activation $h^{1} = W^{1}*x[1\colon t] + b^{1}$ where the weight matrix $W^{1} \in \mathbb{R}^{M_{1}\times t}$ and the bias vector $b^{1} \in \mathbb{R}^{M_1}$. The pre-activation is transformed coordinate by coordinate via a differentiable nonlinear activation function $a^{1}$ to obtain $f^{1} =  a^{1}(h^{1})$.
For every subsequent layer $i,~ 2\leq i \leq L-1$, the output from the previous layer $f^{l-1}$ is transformed to obtain the output of the current layer, $f^{l} = a^{l}(W^{l}*f^{l-1}+b^{l})$, with $W^{l} \in \mathbb{R}^{M_{l}\times M_{l-1}}$ and $b^{l} \in \mathbb{R}^{M_l}$. 
In the final layer, $l=L$, of the neural network the output $\hat{y}$ is obtained as $\hat{y} = a^{L}(W^{L}*f^{L-1}+b^{L})$ where $W^{L} \in \mathbb{R}^{1\times M_{L-1}}$ and $b^{L} \in \mathbb{R}$.
It is well-known that a single hidden layer neural network with an arbitrary number of hidden nodes is a universal approximator \citet{haykin2009neural}. However, in practical scenarios, multi-layer neural networks are adopted. To learn a complex relationship between input and output, we search the space of weight matrices and biases for ideal parameters by optimizing a carefully chosen loss measure on a dataset.

\subsubsection{Some Elements of Statistical Learning Theory}
Statistical Learning Theory \citet{vapnik2013nature} is at the foundation of most machine learning algorithms. A key problem, the discipline addresses is the following. Given a parameterized family of functions $H = \{f(x,\theta), \theta \in \Theta\}$, a joint distribution of random variables $X, Y$, denoted $p(x,y)$ and a loss function $\mathfrak{L}$, the discipline explores the conditions that are necessary and sufficient to solve the optimization problem \begin{equation*}
    \min_{\theta \in \Theta} \mathbb{E}_{p}\left[\mathfrak{L}\left(y,f(x,\theta)\right)\right]
\end{equation*}
through samples generated from the joint distribution $p(x,y)$.\\
In many practical scenarios, one is typically interested in improving the state of the art model performance on benchmark datasets (i.e., to improve $\mathbb{E}_{p}[\mathfrak{L}]$). A probable way to accomplish this task is to expand the search space (i.e. search in $H' \supseteq H$ as 
    $\min_{H'} \mathbb{E}_{p}\left[\mathfrak{L}\left(y,f(x,\theta)\right)\right]\leq \min_{H} \mathbb{E}_{p}\left[\mathfrak{L}\left(y,f(x,\theta)\right)\right]$). 
Many works in the literature \citet{he2015delving,maniatopoulos2021learnable,biswas2021tanhsoft, manessi2018learning} expand the search space through ``extentions" albeit without the explicit mention of the same. In what follows, we formally define ``extentions", state and prove their properties, point out some extensions in the literature and define a few novel extensions and demonstrate their benefits on synthetic as well as real-world datasets.

\section{Analysis}
\label{analysis}
% In this section, we motivate our main ideas and analyse them. We start with the definition of an extension.
In this section, we define extensions and proceed to state and prove their properties. We start with the definition of an extension.
\begin{definition}
An extension of a given function $f: \mathcal{D} \rightarrow \mathbb{R}$ is a function $F: \mathbb{D} \rightarrow \mathbb{R}$ where $\mathcal{D} \subseteq \mathbb{D}$ and $F(x)|_{\{x \in \mathcal{D}\}} = f(x).$ Here $f$ is called as a restriction of $F$.
\end{definition}
\begin{example}
\normalfont
Consider the well-known ReLU given by $f: \mathbb{R} \rightarrow \mathbb{R}$ with 
\begin{equation*}
f(x)=
    \begin{cases}
        x & \text{if } x \geq 0\\
        0 & \text{if } x < 0
    \end{cases}
\end{equation*} 
The PReLU \citet{he2015delving} given by $F: \mathbb{R}^2 \rightarrow \mathbb{R}$ with
\begin{equation*}
F(x, a)=
    \begin{cases}
        x & \text{if } x \geq 0\\
        ax & \text{if } x < 0
    \end{cases}
\end{equation*}
is an extension as it reduces to ReLU when $a=0$ and clearly $\mathbb{R} \subseteq \mathbb{R}^2$
\end{example}
\begin{example}
\normalfont
Both Sigmoid and Swish activations are restrictions of $F: \mathbb{R}^3 \rightarrow \mathbb{R}$ with
\begin{equation*}
F(x, p, q)= \frac{px+q}{1+e^{-x}}
\end{equation*} 
\end{example}
\begin{example}
\normalfont
The SPLASH \citet{tavakoli2021splash} and LeLeLU \citet{maniatopoulos2021learnable} are extensions of ReLU and Leaky ReLU respectively
\end{example}
\begin{example}
\normalfont
Consider $F: \mathbb{R}^4 \rightarrow \mathbb{R}$
\begin{equation*}
F(x, p, q, r)= \frac{px+q}{1+e^{-rx}}
\end{equation*} 
is also an extension of both Sigmoid and Swish activations. The popular Gaussian Error Linear Unit (GELU) \citet{hendrycks2016gaussian} is approximately a restriction of $F(x, p, q, r)$ as
$\text{GELU}(x) \approx \frac{x}{1+e^{-1.702x}} = F(x,1,0,1.702)$. It is also closely related to \citet{bodyanskiy2023learnable}
\end{example}

\subsection{Properties}
In this subsection we state and prove properties of extensions and define our neural network extensions.
\begin{lemma}
Assume that $F: \mathbb{D} \rightarrow \mathbb{R}$ is an extension of a function $f:\mathcal{D} \rightarrow \mathbb{R}$ then
$$\min_{x \in \mathbb{D}} F(x) \leq \min_{x \in \mathcal{D}} f(x)$$
\end{lemma}
\begin{proof}
By definition $F(x) = f(x)$ for $x \in \mathcal{D}$. As a consequence, $\min_{x\in \mathcal{D}} F(x) = \min_{x\in \mathcal{D}} f(x)$. Again by definition $\mathcal{D} \subseteq \mathbb{D}$ and on a larger set minimum value can only decrease. So, 
$\min_{x \in \mathbb{D}} F(x) \leq \min_{x \in \mathcal{D}} f(x).$ \\
\end{proof}
\begin{lemma}\label{transitivity}
Suppose $F: \mathbb{D} \rightarrow \mathbb{R}$ is an extension of $f:\mathcal{D} \rightarrow \mathbb{R}$ and $G:\mathfrak{D} \rightarrow \mathbb{R}$ is an extension of $F$. Then $G$ is also an extension of $f$.
\end{lemma}
\begin{proof}
Since $F$ is an extension of $f$ and $G$ is an extension of $F$, we have $\mathcal{D}\subseteq\mathbb{D}$ and $\mathbb{D}\subseteq\mathfrak{D}$. So $\mathcal{D}\subseteq\mathfrak{D}$. Similarly, from the definition we have $F(x)|_{x\in \mathcal{D}} = f(x)$ and $G(x)|_{x\in \mathbb{D}} = F(x)$. As $\mathcal{D}\subseteq\mathbb{D}$, we have $G(x)|_{x\in \mathcal{D}}=F(x)=f(x)$. So $G$ is an extension of $f$. Moreover, the relation extension between two functions is transitive. \\
\end{proof} 
\begin{lemma}\label{anti-symmetry}
Suppose $F: \mathbb{D} \rightarrow \mathbb{R}$ is an extension of $f:\mathcal{D} \rightarrow \mathbb{R}$ and $f:\mathcal{D} \rightarrow \mathbb{R}$ is an extension of $F$ as well. Then $f=F$.
\end{lemma}
\begin{proof}
It is given that $f$ is an extension of $F$, so $\mathcal{D} \subseteq \mathbb{D}$ and $f(x)=F(x)$ on $\mathcal{D}$. It is also given that $F$ is an extension of $f$, so
$\mathbb{D} \subseteq \mathcal{D}$ and $F(x)=f(x)$ on $\mathbb{D}$. So we have $\mathcal{D} = \mathbb{D}$ and $f(x)=F(x)$ and the relation extension is anti-symmetric.
\end{proof}
\begin{lemma}
The relation extension forms a partial order on the space of functions.
\end{lemma}
\begin{proof}
From Lemma \ref{transitivity} and Lemma \ref{anti-symmetry}, the relation extension is anti-symmetric and transitive. A function $f$ is, by definition, an extension of itself. So the relation extension is reflexive as well. Hence the relation extension forms a partial order on the space of functions. \\
\end{proof}
\begin{lemma}
Let $d$ be the number of parameters. Consider a feed forward neural network $\mathsf{N}$ of $l$ hidden layers with the corresponding activations $a_1, a_2, \cdots, a_l$. Typically $a_i \in \mathcal{S} \coloneqq \{\text{ReLU, GELU, Tanh, Sigmoid, ... }\}$, a library of activation functions and is common for each node of the network. The neural network $\mathsf{N}$ is a function $\mathsf{N}: \mathbb{R}^{d} \times [|\mathcal{S}|]^{l} \rightarrow \mathbb{R}$. Let $a=(a_1, a_2, \cdots, a_{|\mathcal{S}|})$ be the vector of activation functions. The $l$ hidden layer neural network $\mathsf{M}$ with the activation functions $g_1(a), g_2(a), \cdots, g_l(a)$, where $g_i = \lambda_i^{T} a$ with $\lambda_i \in \mathbb{R}^{|\mathcal{S}|}$  is an extension of $\mathsf{N}$ . 
\end{lemma}
\begin{proof}
Recall that $d$, denotes the number of parameters of the network. The neural network $\mathsf{M}$ is a function of ${d+|\mathcal{S}|l}$ parameters, i.e., $\mathsf{M}: \mathbb{R}^{d}\times \mathbb{R}^{|\mathcal{S}|l} \rightarrow \mathbb{R}$ and clearly $\mathbb{R}^{d} \times [|\mathcal{S}|]^{l} \subseteq \mathbb{R}^{d}\times \mathbb{R}^{|\mathcal{S}|l}$. \\ 
Note that $g_i = a_i$ if $\lambda_i = (0,0,\cdots, 1, \cdots, 0)$ where $1$ is at position $i$. Hence $\mathsf{M}$ reduces to $\mathsf{N}$ for these choices of $\lambda_i$ i.e., $\mathsf{M}$ is an extension of $\mathsf{N}$. \\
\end{proof}
\begin{corollary}
    The feed forward neural network $\mathsf{Q}$ with $l$ hidden layers and the corresponding activation functions $g_i$ given by $g_i = a^T \Lambda_i a + \lambda_i^{T} a$, where $\Lambda_i$ is an upper triangular $|\mathcal{S}| \times |\mathcal{S}|$ parameter matrix, is an extension of $\mathsf{N}.$
\end{corollary}
\begin{proof}
Follows from Lemma \ref{transitivity} as $g_i = \lambda^T_{i}a$  if $\Lambda_i = \textbf{0}$. \\
\end{proof}

\begin{remarks}
Extensions of neural networks are not unique. 
\end{remarks}
\begin{remarks}
The extension relation forms a partial order in the space of neural networks and extensions dominate restrictions on a given dataset/data distribution.
\end{remarks}
\begin{remarks}
Imposing constraints for e.g. $\overrightarrow{1}^T\lambda_i=1$ or $\lambda_i \succcurlyeq \overrightarrow{0}$ still ensures that $\mathsf{M}$ is an extension of $\mathsf{N}$ and is related to \citet{manessi2018learning}.   
\end{remarks}
\begin{remarks}
In general $\mathsf{Q}$ and $\mathsf{M}$ have better performance compared to $\mathsf{N}$ for a given machine learning task.
\end{remarks}
\begin{remarks}
A reason for performance improvement mentioned in many works in the literature, for e.g.\citet{he2015delving,maniatopoulos2021learnable,biswas2021tanhsoft, manessi2018learning}, is the construction of 
$H' \supseteq H$ through extensions of neural networks.

\end{remarks}

\subsection{Learnable Activations}
We call the activations given by
\begin{equation} \label{LLA} g_i = \lambda_i^{T} a.\end{equation}
\begin{equation} \label{QLA} g_i = a^T \Lambda_i a + \lambda_i^{T} a. \end{equation}

as Linear Learnable Activation (LLA) and Quadratic Learnable Activation (QLA) respectively
\begin{remarks}
The choice of learnable activation (LLA or QLA) in a neural network is not hyper parameter optimization. We are not choosing the best activation functions for the neural network. The hyper parameter optimization in this context is the choice of the activation library $\mathcal{S}$.
\end{remarks}
\begin{remarks}
To ensure that the convergence properties of $\mathsf{M}$ are similar to $\mathsf{N}$,
constraints - i.e., $\overrightarrow{1}^T\lambda_i=1$ and $\lambda_i \succcurlyeq \overrightarrow{0}$ are enforced.
\end{remarks}
\begin{remarks}
As the extensions get complicated, the optimization process complexity and the number of parameters increase    
\end{remarks}
\subsubsection{\textbf{Time and Space Complexity}}
\label{complexity}
Given the library of activation functions $\mathcal{S} \coloneqq \{\text{ReLU, GELU, Tanh, Sigmoid, ... }\}$ and  per iteration computation cost, $C$, of $\mathsf{N}$, the per iteration computation cost of $\mathsf{M}$ with LLA is $O(|\mathcal{S}|)C$. Under the assumption that the number of parameters per layer in $\mathsf{N}$ is $P$, the number of parameters in $\mathsf{M}$ with LLA is $P+|\mathcal{S}|$. \\
Similarly, in the case of  $\mathsf{M}$ with QLA the per iteration computation cost is $O(|\mathcal{S}|)C+ O(|\mathcal{S}|^2)$ and the number of parameters per layer in $\mathsf{M}$ with QLA is $P+O(|\mathcal{S}|^2)$

\section{Experiments}
We evaluate our extensions on eight test functions taken 
from \url{http://www.sfu.ca/~ssurjano/index.html} and defined in Table \ref{test_functions}.
\begin{table*}[!htbp]
\small
\centering
  \caption{Test Functions for Evaluation}
  \label{test_functions}
  \begin{tabular}{ccl}
    \toprule
    Name & Definition\\
    \midrule
    Ackley & $f(x)=-20 \exp\left({-0.2*\sqrt{\frac{1}{d}\displaystyle\sum^{d}_{i=1}x_i^2}}\right)-\exp\left({\frac{1}{d}\displaystyle\sum^d_{i=1}\cos{2\pi x_i}}\right) +20+\exp(1)$ \\ 
    \hline
    Shubert & $f(x)= \displaystyle\prod^d_{k=1}\left(\displaystyle\sum^5_{i=1} i \cos((i+1)x_k+i)\right)$  \\
    \hline
    Hyper Ellipsoid & $f(x)=\displaystyle\sum^{d}_{i=1}\displaystyle\sum^{i}_{j=1} x_{j}^{2}$ \\
    \hline
    Levy & $f(x)=\sin^{2}(\pi w_{1})+\displaystyle\sum^{d}_{i=1}(w_i-1)^2[1+10\sin^2(\pi w_{i}+1)] + (w_d-1)^2[1+\sin^2(2\pi w_d)]$  \\
     & where $w_i = 1+\frac{(x_{i}-1)}{4}$ for all $i=1, \cdots, d$  \\
    \hline
    Styblinski & $f(x)=\frac{1}{2}\displaystyle\sum^{d}_{i=1}\left(x^4_{i}-16x^2_{i}+5x_{i}\right)$ \\
    \hline
    Shekel & $f(x)=-\displaystyle\sum^{m}_{i=1}\left(\displaystyle\sum^{4}_{j=1}(x_j-C_{ji})^2 + \beta_i \right)^{-1}$ \\ 
    & $C$ and $\beta$ are custom parameters\\ 
    \hline
    
%     Shekel & $f(x)=-\sum^{m}_{i=1}\left(\sum^{4}_{j=1}(x_j-C_{ji})^2 + \beta_i \right)^{-1}$ & Defined only for $d=4$, $m=10$,\\
%     &   & $\beta = \frac{1}{10}(1,2,2,4,4,6,3,7,5,5)$\\
%     &   & $C = \begingroup % keep the change local
% \setlength\arraycolsep{2pt}
% \begin{pmatrix}  4 & 1 & 8 & 6 & 3 & 2 & 5 & 8 & 6 & 7 \\
%                                4 & 1 & 8 & 6 & 7 & 9 & 3 & 1 & 2 & 3.6 \\
%                                4 & 1 & 8 & 6 & 3 & 2 & 5 & 8 & 6 & 7 \\
%                                4 & 1 & 8 & 6 & 7 & 9 & 3 & 1 & 2 & 3.6 \\
%     \end{pmatrix} \endgroup$\\
%     \hline

    % Drop Wave & $f(x)=-\frac{1+\cos\left(12\sqrt{\|x\|^{2}}\right)}{0.5\|x\|^2 + 2}$  \\
    % \hline

    Griewank & $f(x)=\displaystyle\sum^{d}_{i=1}\frac{x_i^2}{4000} - \Pi^{d}_{i=1}\cos\left(\frac{x_i}{\sqrt{i}}\right) + 1$  \\
    % \hline
    % Rosenbrock & $f(x)=\sum^{d-1}_{i=1}\left[100(x_{i+1}-x^2_{i})^2+(x_i-1)^2\right]$  \\ 
    \hline
    Zhou & $f(x)=\frac{10^d}{2}\left[\phi(10(x-1/3))+\phi(10(x-2/3))\right]$ \\
         & where $\phi(x)=(2\pi)^{-d/2}\exp(-0.5\|x\|^2)$  \\
    % \hline   
    % Zakharov & $f(x)=\sum^{d}_{i=1}x_i^2+\left(\sum^{d}_{i=1}0.5ix_i\right)^2+\left(\sum^{d}_{i=1}0.5ix_i\right)^4$ \\
    \bottomrule
  \end{tabular}
\end{table*}
All these test functions are rather complex and pose significant difficulty for the learning process especially for vanilla activation functions. \\ 

Our experimental setting is as follows. For all these test functions our experimental evaluation is in dimension $d=2$. We fixed the seed to be $13$. We choose a simple feed-forward neural network with $2$ input nodes, $2$ hidden layers with $20$ hidden nodes each and $1$ output node with default initialization of weights and biases as described in \url{https://pytorch.org/docs/stable/generated/torch.nn.Linear.html}.

For LLA and QLA the corresponding parameters are initialized with Kaiming normal initialization as described in \url{https://pytorch.org/docs/stable/nn.init.html#torch.nn.init.kaiming_normal_} and the library chosen is $\mathcal{S} =\{\text{ReLU, GELU, Tanh, Sigmoid}\}$ \\

We set the learning rate as $0.001$, and ran the experiment for each test function for $500$ epochs with a batch size of $256$. Our loss function is Mean Squared Error. We have chosen the Adam optimizer for the learning process and a dataset of $15000$ points for each test function is generated through a quasi montecarlo process that is described in  \url{https://docs.scipy.org/doc/scipy/reference/generated/scipy.stats.qmc.Halton.html}. We measure the performance in terms of Mean Absolute Error (MAE) and Mean Squared Error (MSE) and summarize in Table \ref{metrics1}

\begin{table*}[!htbp]
\tiny
  \centering
  \caption{Comparison of LLA and QLA against vanilla activations on test functions.}
  \label{metrics1}
  \begin{tabular}{ccccccl}
    \toprule
    Name & ReLU & GELU & Tanh & Sigmoid &  LLA & QLA\\
         & MSE\textbar MAE &MSE\textbar MAE &MSE\textbar MAE & MSE\textbar MAE & MSE\textbar MAE & MSE\textbar MAE \\
    \midrule
    Ackley &0.42\textbar0.495 &0.389\textbar0.465 &0.358\textbar\textbf{0.449} &0.367\textbar0.452 &\textbf{0.335}\textbar0.469 & 0.357\textbar0.454\\
    \hline
    Shubert &1.552\textbar0.818 &1.581\textbar0.838 &1.664\textbar0.855 &2.16\textbar0.915 &1.432\textbar0.776 &\textbf{0.345}\textbar\textbf{0.437}\\ 
    \hline
    Hyper  & & & & & \\
    Ellipsoid &3881.647\textbar47.877 &3015.138\textbar44.654 &24875495.187\textbar3998.474 &24925986.568\textbar4003.899 &4326.411\textbar50.334 &\textbf{19.361}\textbar\textbf{3.224} \\
    \hline
    Levy & 3.717\textbar1.219 & 2.316\textbar1.046 &2.107\textbar1.0 &3.403\textbar1.354 &\textbf{1.386}\textbar\textbf{0.861} &1.99\textbar0.994 \\
    \hline
    Styblinski &2.386\textbar1.157 &0.105\textbar0.224 &15.954\textbar1.369 &355.686\textbar12.665 &1.085\textbar 0.38 &\textbf{0.075}\textbar\textbf{0.195}\\
    \hline
    Shekel &1.581\textbar0.925 &1.415\textbar0.856 &\textbf{0.109}\textbar\textbf{0.236} &1.905\textbar1.041 &0.649\textbar0.562 &0.123\textbar0.261 \\
    \hline
    % Drop Wave &0.012\textbar0.08 &0.012\textbar0.08 &0.012\textbar0.08 &0.012\textbar0.08 &0.012\textbar0.081 & 0.012\textbar0.081 \\
    % \hline
    Griewank &1.498\textbar0.9 & 1.811\textbar1.009 &5.635\textbar1.672 &3.178\textbar1.247 &\textbf{1.019}\textbar\textbf{0.754} &426.561\textbar15.083 \\
    % \hline
    % Rosenbrock &740602430.0\textbar &3163387400.0\textbar &83762545000.0\textbar &83762680000.0\textbar &161812740.0\textbar &\textbf{31142.695}\textbar \\
    % &12325.426 &30221.994 &143072.6 &143072.97 &5424.28 &\textbf{128.014}\\
    \hline
    Zhou &0.012\textbar0.076 &0.061\textbar0.174 &0.011\textbar0.077 &1.042\textbar0.57 &0.006\textbar0.062 &\textbf{0.003}\textbar\textbf{0.042}\\
    % \hline
    % Zakharov &42037.971\textbar104.207 &27052.957\textbar66.979 &84509169.867\textbar4157.495 &84546755.209\textbar4175.05 &60902.571\textbar150.91 & \textbf{150.982}\textbar\textbf{7.738} \\
    \bottomrule
  \end{tabular}
\end{table*}

\begin{figure}[h]
        \begin{subfigure}[b]{0.5\textwidth}
                \includegraphics[height=\linewidth, width=\linewidth]{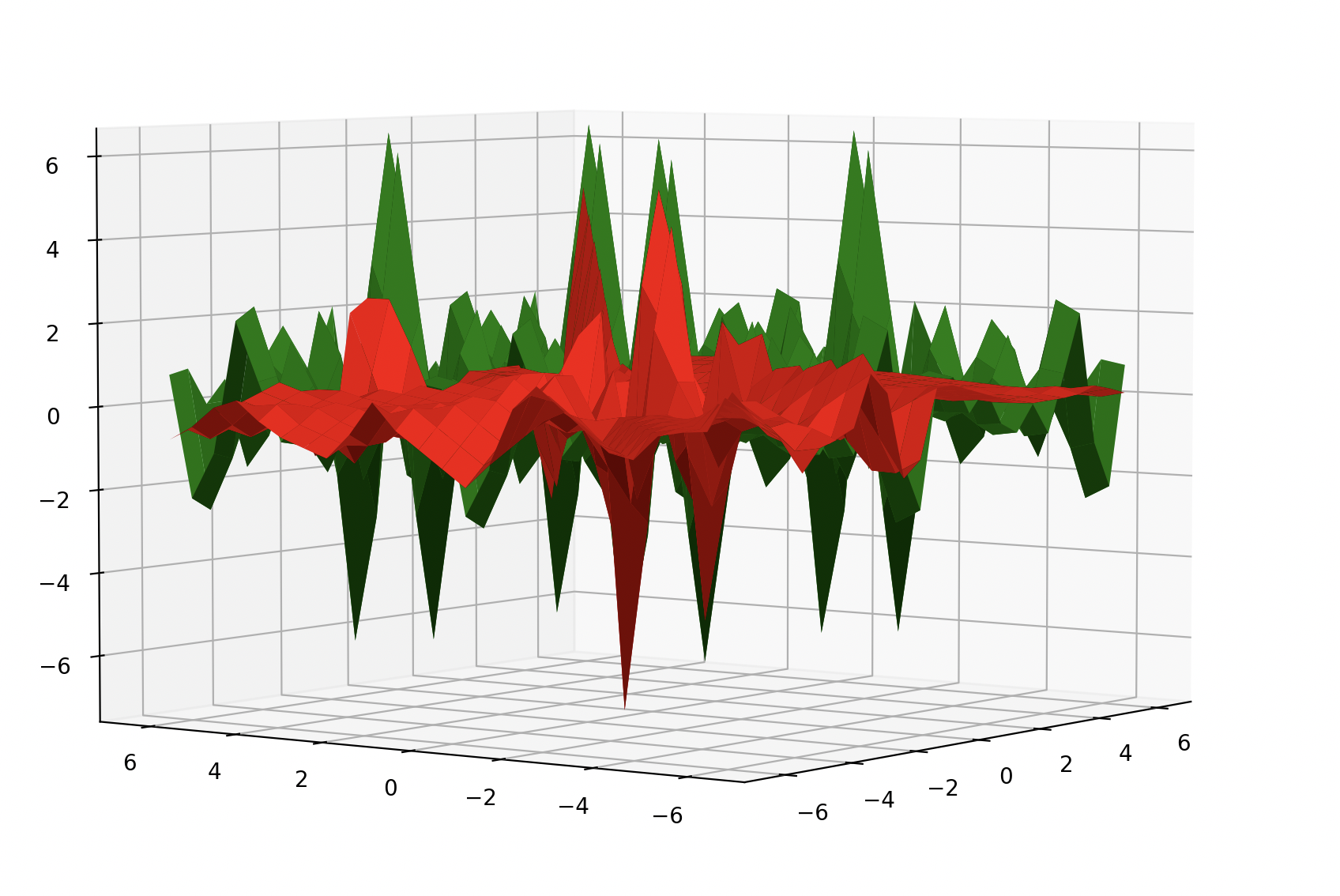}
                \caption{Activation is ReLU}
                \label{shubert_relu2}
        \end{subfigure}%
        \begin{subfigure}[b]{0.5\textwidth}
                \includegraphics[height=\linewidth, width=\linewidth]{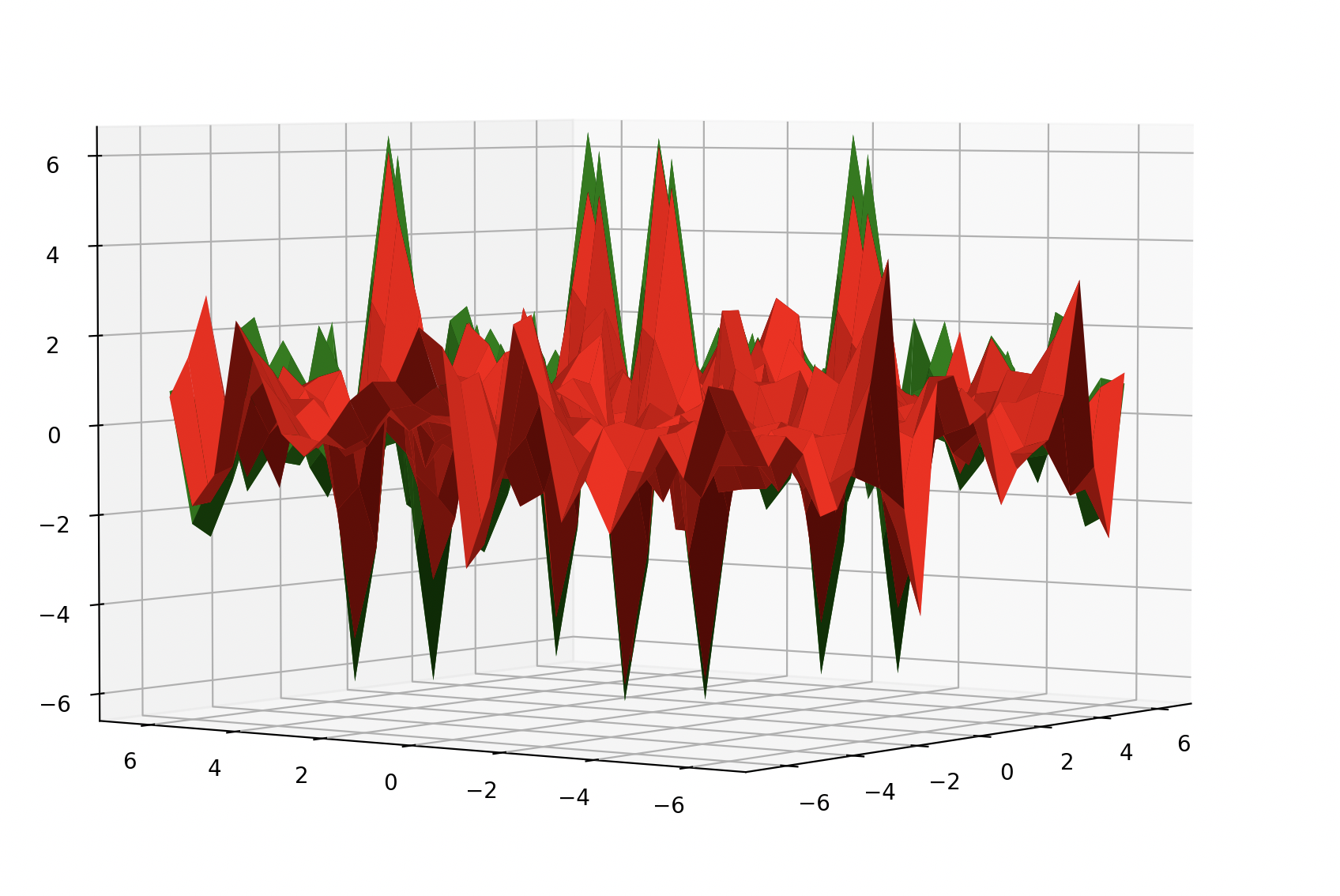}
                \caption{Activation is QLA}
                \label{shubert_qla3}
        \end{subfigure}%
        \caption{Visual comparison between ReLU and QLA for Shubert function on test dataset.}\label{visual_comp}
\end{figure}

It is easily seen that in most cases LLA and QLA are superior to individual activations and QLA is superior to LLA, in alignment with the analysis of Section \ref{analysis}. The differences in the learning process are noticed visually as well. For example, observe the differences between learned surfaces for Shubert function on a grid of test data in the case of ReLU vs QLA shown in the Figure \ref{visual_comp}. 

\begin{figure}[h]
\centering
    \includegraphics[width=\linewidth, scale=1.3, trim={0 5mm 0 0}, clip]{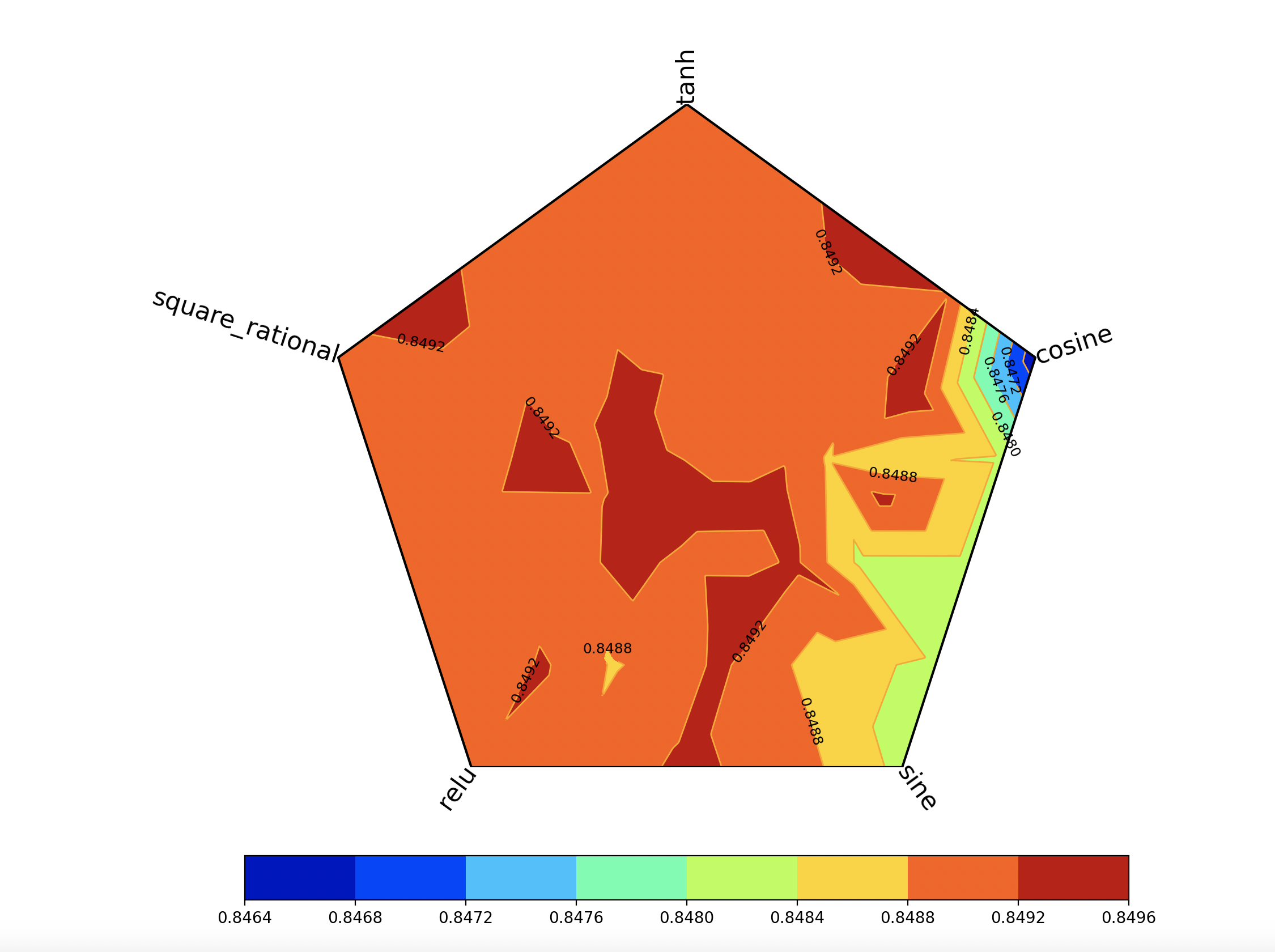}
    \caption{Depiction of level sets of MSE on the pentagon for Shubert test function}
    \label{baryplot}
\end{figure}

\begin{figure}[h]
        \centering
        \begin{subfigure}[b]{0.3\textwidth}
            \centering
            \includegraphics[height=0.7\linewidth,width=\linewidth]{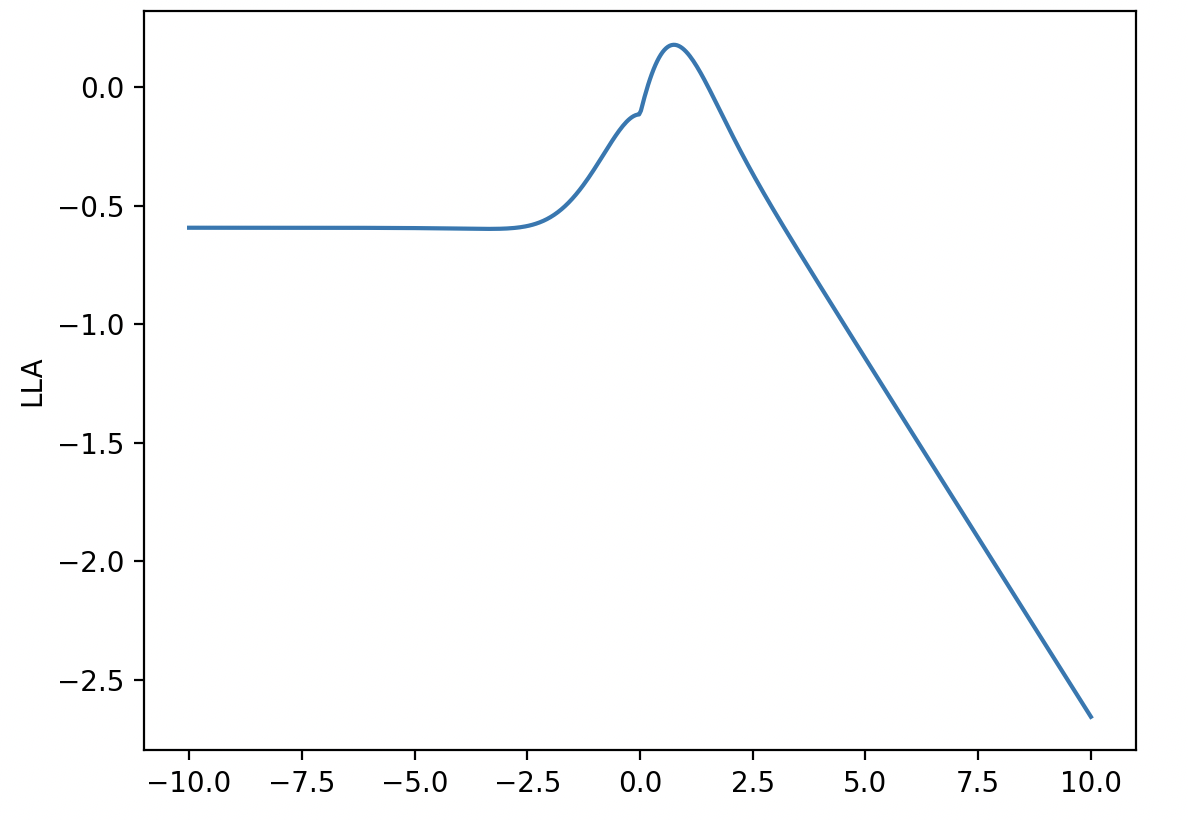}
            \caption{\footnotesize Activation in $1^{\text{st}}$ hidden layer  with LLA}%  
            \label{LLA_layer1}
        \end{subfigure}
        % \hfill
        \begin{subfigure}[b]{0.3\textwidth}  
            \centering 
            \includegraphics[height=0.7\linewidth, width=\linewidth]{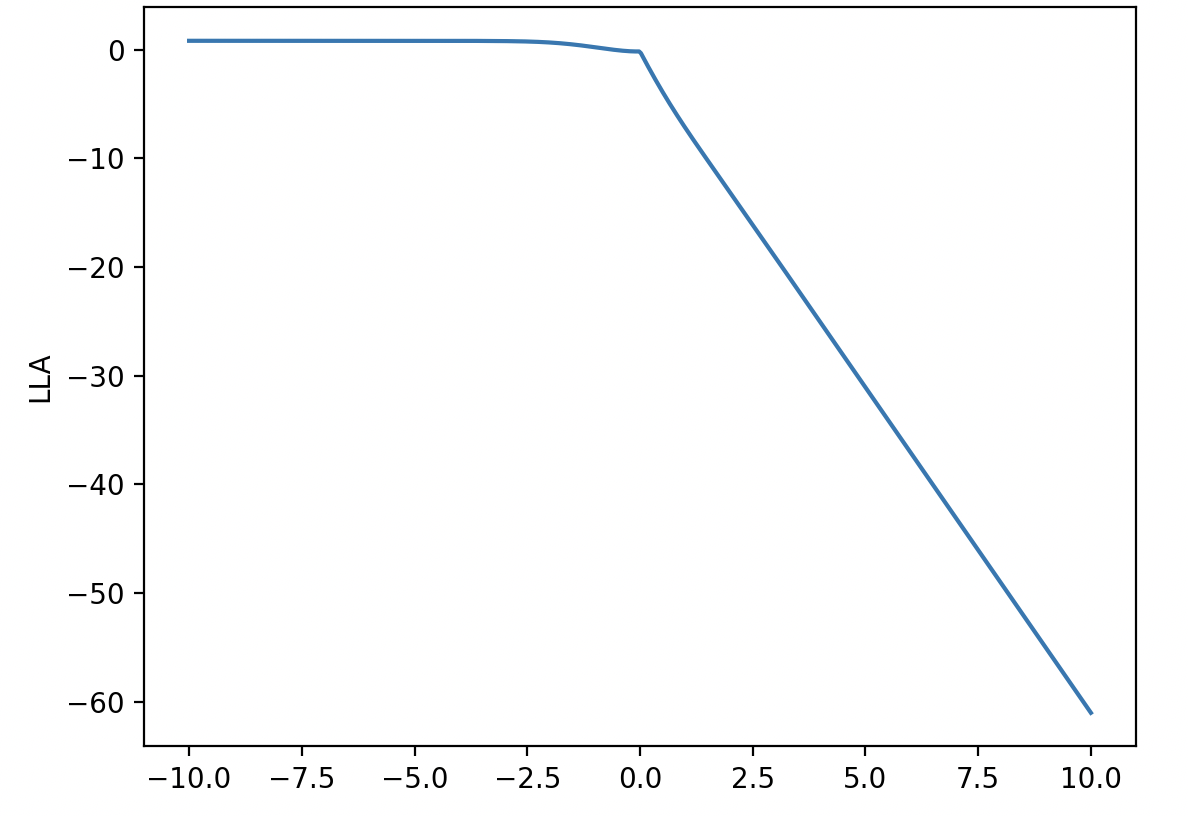}
            \caption{\footnotesize Activation in $2^{\text{nd}}$ hidden layer with LLA}%  
            \label{LLA_layer2}
        \end{subfigure}
        \vskip\baselineskip
        \begin{subfigure}[b]{0.3\textwidth}   
            \centering 
            \includegraphics[height=0.7\linewidth,width=\linewidth]{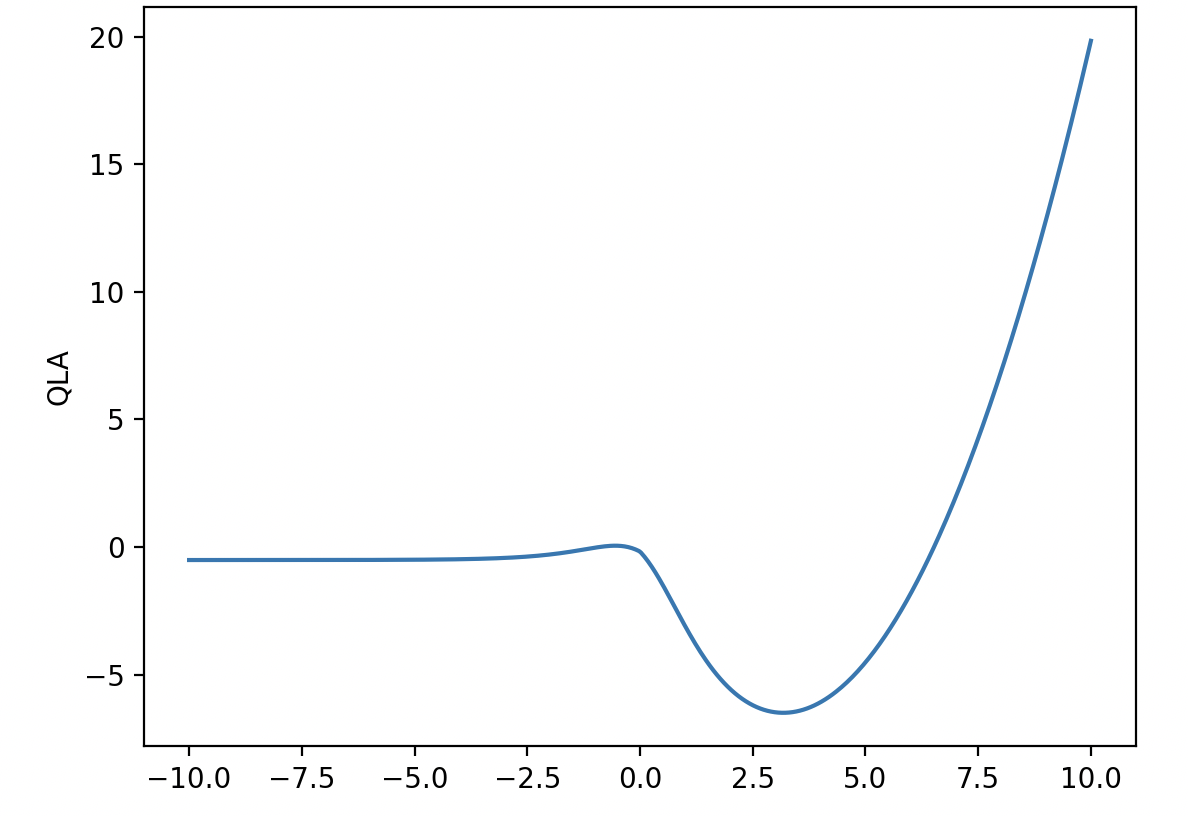}
            \caption{\footnotesize Activation in $1^{\text{st}}$ hidden layer with QLA}%   
            \label{QLA_layer1}
        \end{subfigure}
        % \hfill
        \begin{subfigure}[b]{0.3\textwidth}   
            \centering 
            \includegraphics[height=0.7\linewidth,width=\linewidth]{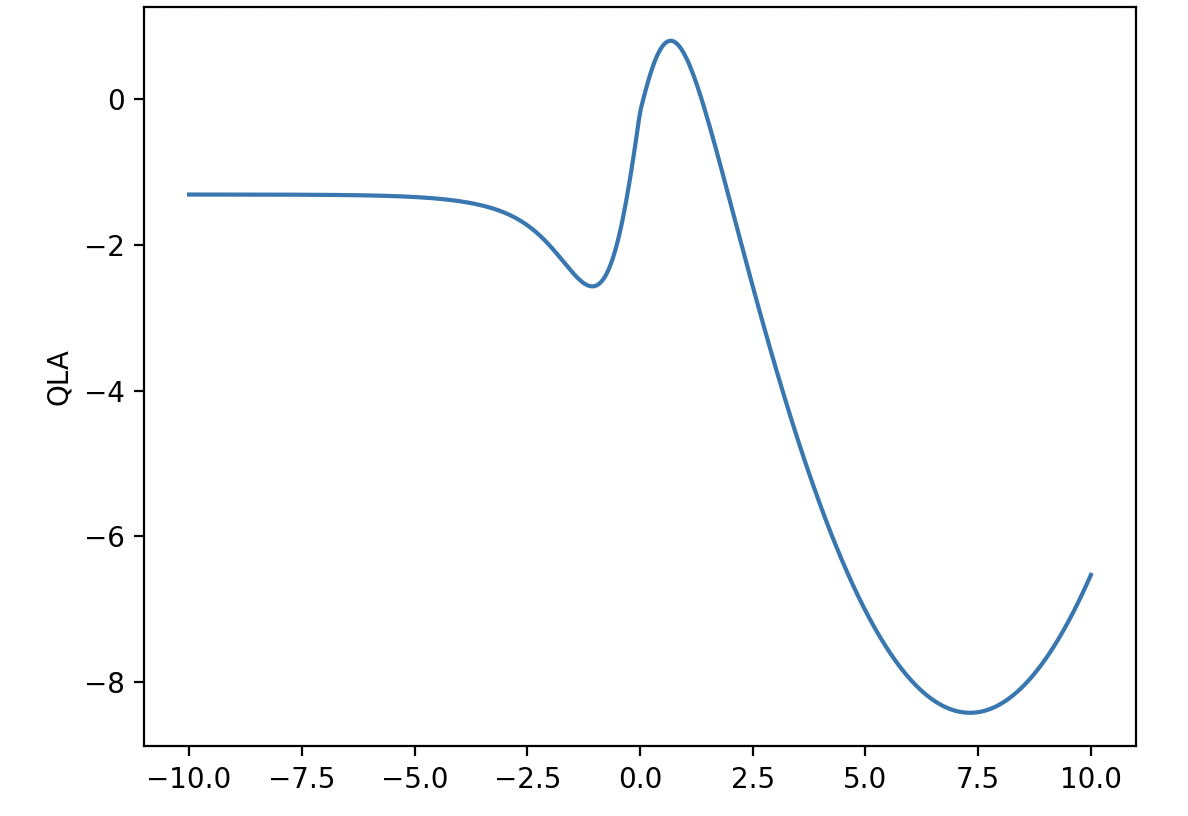}
            \caption{\footnotesize Activation in $2^{\text{nd}}$ hidden layer with QLA}%
            \label{QLA_layer2}
        \end{subfigure}
        \caption{Plot of activations learned for the Shubert function in hidden layers $1$ and $2$ for the network under consideration.} 
        \label{learned_activations}
    \end{figure}

The green background is the actual surface of Shubert function and the red surfaces are learned with the corresponding choice of activations. It is easily seen that QLA generates a better fit mostly due to non-trivial activations that are learned compared to ReLU, a piecewise linear activation.
Further appreciation is gained by comparing the plots of learned activation functions in the case of LLA and QLA configuration, shown in Figure\ref{learned_activations}, against ReLU, GELU, Tanh and Sigmoid. The learned activation functions appear to adapt to the intricacies of the surface of the test function, especially in the case of QLA. \\
We note here that, for easy visualization purposes, experiments are conducted on synthetic test functions in dimension $2$. We noticed similar phenomena in higher dimensions as well, also evident from experiments on real-world datasets described below. \\
Another experiment that provides further insights is the following. We consider the same network as described above for the Schubert function in dimension $2$. The choice of library of functions is $\mathcal{S} =\{\text{ReLU, Tanh, Sine, Cosine, Square\_rational}\}$ where $\text{Square\_rational}(x) = \frac{x^2}{1+x^2}$. We choose the LLA configuration for both hidden layer activations with initialization from barycentric coordinates of a pentagon (i.e., $\lambda_1 = \lambda_2 = \lambda$ chosen from the interior of a pentagon). We optimize the network on the rest of the parameters (only $\lambda$ is fixed) over $200$ epochs and plot the mean squared error as shown in Figure \ref{baryplot}. The colour indicates the level sets of MSE. 
We conclude that learnable activations (LLA and QLA) 
adapt to learn complex relationship between inputs and outputs and boost the performance of the base networks.
\subsubsection*{Issues with LLA and QLA}
\begin{itemize}
    \item Both LLA and QLA are sensitive to the initialization of the learning process. We noticed both performance improvement and degradation based on the choice of initialization compared to metrics mentioned in Table \ref{metrics1}.
    \item Cardinality of the library $\mathcal{S}$ adds to the complexity of the learning process (see \ref{complexity}).
    \item The choice of elements of library $\mathcal{S}$ is application dependent. E.g., we choose functions $\sin(x)$ and $\cos(x)$ as elements of the library to exploit the periodicity in the time series datasets in the subsequent experiments.
\end{itemize}

\subsection{Experiments on real-world time series datasets}
We choose four real-world time series benchmark datasets ETTh1, ETTh2, ETTm1 and ETTm2 available at
\url{https://github.com/zhouhaoyi/ETDataset/tree/main/ETT-small}. ETTh1 and ETTh2 have measurements at hourly frequency, while ETTm1 and ETTm2 have measurements at $15$ min frequency. Each dataset has timestamped measurements of $7$ features of electricity transformers namely, HUFL (High UseFul Load), HULL (High UseLess Load),	MUFL (Middle UseFul Load), MULL (Middle UseLess Load), LUFL(Low UseFul Load), LULL (Low UseLess Load) and OT (Oil Temperature).

In the case of each dataset, given $512$ historical points of these features, the task is to forecast $96$ points into the future for all the features.
For the forecasting task, our network consists of $512\times 7$ dimensional input layers. $96\times 7$ dimensional output layer and $2$ hidden layers of dimension $7$ each. We choose QLA configuration for activations with $\mathcal{S} =\{\text{ReLU, GELU, Sine, Cosine}\}$ as the library. For the reproducibility of results we set the seed as $36$ for all the experiments. Our initialization for weights and biases is default and for QLA parameters it is Kaiming normal as described in the earlier experiment for synthetic test functions. Each dataset spans over $2$ years. We have chosen the last $4$ months for the test dataset. The training and validation datasets comprise of first $16$ months and $16-20$ months of data and the dataset is normalised for the learning process. We chose MSE as the error metric. We compare QLA against vanilla ReLU activation for these datasets. 
The performance metrics are summarized in the Table \ref{ettmetrics}
\begin{table}[h]
\centering
\begin{tabular}{lll}
\hline
Dataset & MSE-ReLU & MSE-QLA        \\ \hline
ETTh1   & 1.561    & \textbf{0.953} \\ \hline
ETTm1   & 0.597    & \textbf{0.540} \\ \hline
ETTh2   & 0.717    & \textbf{0.563} \\ \hline
ETTm2   & \textbf{0.203 }   & 0.208          \\ \hline
\end{tabular}
\caption{Performance metrics on ETT Datasets}
\label{ettmetrics}
\end{table}
Based on the metrics in Table \ref{ettmetrics}, it is easily seen that QLA outperforms or on par with ReLU on the chosen time series datasets.
Also, as is evident from Figure \ref{baryplot}, optimization of neural networks under LLA/QLA is extremely non-convex. Hence choice of initial point, stopping criterion etc., have a significant role in the optimization process. For example,  we have observed it in the case of ETTm2 dataset, on introducing an early stopping criterion based on validation dataset, we observe that MSE of QLA drops to $0.197$ surpassing ReLU. \\

\begin{figure}[htbp]
        \centering
        \begin{subfigure}[b]{0.5\textwidth}
            \centering
            \includegraphics[height=0.23\linewidth, width=\linewidth]{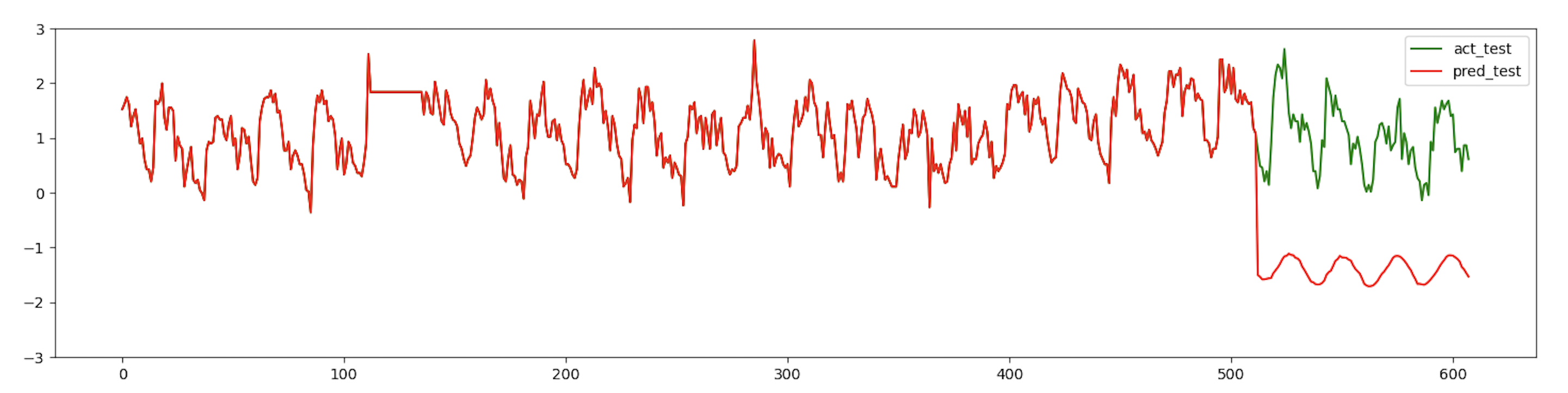}
            \caption{Actual vs prediction with ReLU}%  
        \end{subfigure}
        % \hfill
        \vskip\baselineskip
        \begin{subfigure}[b]{0.5\textwidth}  
            \centering 
            \includegraphics[height=0.23\linewidth, width=\linewidth]{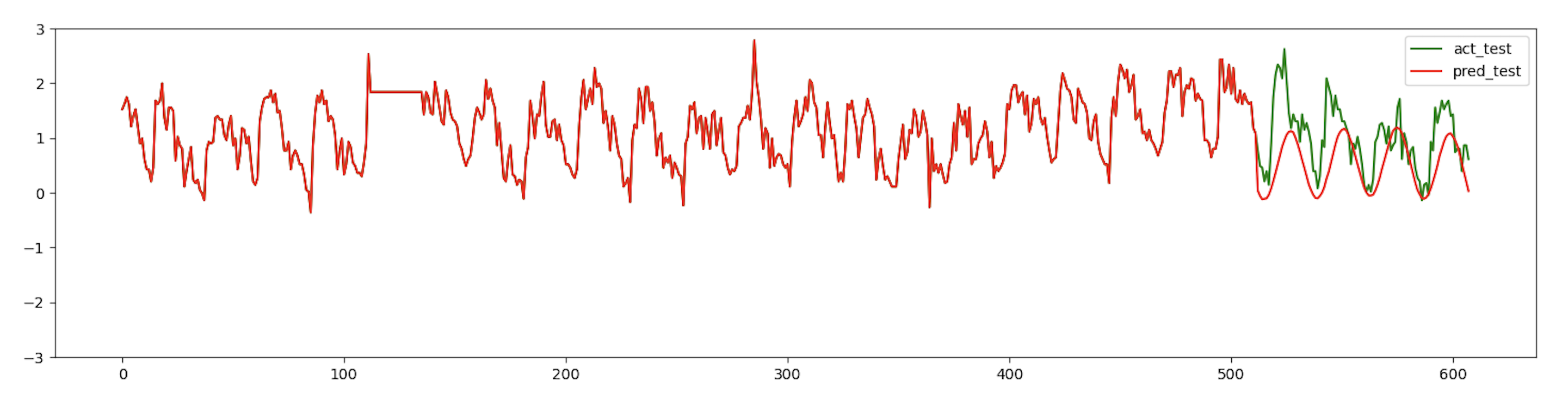}
            \caption{Actual vs prediction with QLA}%  
        \end{subfigure}
        \caption{Actual vs forecast for typical test data sample point of HULL feature of ETTh1 dataset.} 
        \label{testsample_plot}
    \end{figure}

One of the reasons for this improvement is the exploitation of periodicity in the datasets. Consider the plots shown in Figure \ref{testsample_plot} for a typical sample point of HULL in ETTh1 test dataset. Green plot is the ground truth, the last $96$ points of the red plot are the predictions and the first $512$ points form the history utilized for forecasting. It is evident that QLA models periodicity in the data much better than ReLU most likely due to the presence of $\sin(x)$ and $\cos(x)$ in the library.

% \newpage
% \newpage
% Points to expand
% \begin{itemize}
%     \item Choice of initial point for learnable activation
%     \item Cubic extension
%     \item Other extensions in the literature
%     \item how the activation functions look like in general and overtime ---Done
% \end{itemize}
\section{Conclusion and Future Work}
We have shown that expansion of the search space through extensions is a framework that boosts the performance of neural networks.
Based on the framework we proposed LLA and QLA that are extensions and analysed their performance on synthetic as well as real world datasets. Similar to LLA and QLA, exploring higher-order extensions like cubic learnable activation, node specific learnable activations, unlike layer specific activations in this work, are promising directions for future work. We noticed encouraging benefits in the case of cubic learnable activation, however, the optimization process faces convergence issues and potentially requires additional constraints on the elements of library $\mathcal{S}$.
Initialization strategies, based on the elements of the library $\mathcal{S}$, for learnable activation parameters to avoid local optima and applications in relevant domains are other promising future directions.

\bibliography{LearnableActivation/components/refs}
\bibliographystyle{iclr2024_conference}

% \appendix
% \section{Appendix}

\end{document}